\documentclass[conference]{IEEEtran}
\IEEEoverridecommandlockouts

\usepackage{thmbox} 
\newtheorem[M, bodystyle=\normalfont\noindent]{definition}{Definition}

\newtheorem[M, bodystyle=\normalfont\noindent]{problem}{Problem}

\usepackage[]{algorithmic} 
\usepackage{amssymb}

\usepackage{cite}
\usepackage{amsmath,amssymb,amsfonts}
\usepackage{algorithmic}
\usepackage{graphicx}
\usepackage{textcomp}
\usepackage{xcolor}
\usepackage[utf8]{inputenc}
\usepackage{newunicodechar}
\newunicodechar{−}{\textminus}
\usepackage{bm}
\usepackage{tabularx} 

\usepackage{mdwlist}
\usepackage{algorithm}
\usepackage{algorithmic}
\usepackage{booktabs}
\usepackage{color}

\usepackage{amsthm,amssymb}
\usepackage{amsmath}
\usepackage[font=small]{caption}
\usepackage[skip=3pt]{subcaption}
\usepackage{multirow}
\usepackage{mathrsfs}
\usepackage{amsbsy}
\usepackage[mathscr]{eucal} 
\usepackage[flushleft]{threeparttable}
\usepackage[hidelinks]{hyperref}
\usepackage{comment}
\usepackage{paralist}
\usepackage{pifont}

\usepackage{bbding} 

\usepackage{wrapfig}

\def\BibTeX{{\rm B\kern-.05em{\sc i\kern-.025em b}\kern-.08em
    T\kern-.1667em\lower.7ex\hbox{E}\kern-.125emX}}

\begin{document}

\title{Homotopy-Guided Self-Supervised Learning of Parametric Solutions for AC Optimal Power Flow
}

\author{Shimiao~Li$^{1}$, Aaron~Tuor$^{2}$, 
Draguna~Vrabie$^{2}$, Larry~Pileggi$^{3}$, Jan~Drgona$^{4}$\\
\thanks{Authors are with $^1$University at Buffalo; $^2$Pacific Northwestern National Lab;
$^3$Carnegie Mellon University; 
$^4$John Hopkins University; 
}
}

\maketitle

\begin{abstract}
Learning to optimize (L2O) parametric approximations of AC optimal power flow (AC-OPF) solutions offers the potential for fast, reusable decision-making in real-time power system operations. However, the inherent nonconvexity of AC-OPF results in challenging optimization landscapes, and standard learning approaches often fail to converge to feasible, high-quality solutions. This work introduces a \textit{homotopy-guided self-supervised L2O method} for parametric AC-OPF problems. The key idea is to construct a continuous deformation of the objective and constraints during training, beginning from a relaxed problem with a broad basin of attraction and gradually transforming it toward the original problem. The resulting learning process improves convergence stability and promotes feasibility without requiring labeled optimal solutions or external solvers. We evaluate the proposed method on standard IEEE AC-OPF benchmarks and show that homotopy-guided L2O significantly increases feasibility rates compared to non-homotopy baselines, while achieving objective values comparable to full OPF solvers. These findings demonstrate the promise of homotopy-based heuristics for scalable, constraint-aware L2O in power system optimization.
\end{abstract}


\section{Introduction}\label{sec: intro}

Learning to Optimize (L2O)~\cite{vanhentenryck2025} offers an emerging alternative to repeatedly solving large-scale optimization problems in power system operations. Instead of running a numerical solver from scratch for each new operating condition, the idea is to learn a parametric solution map that directly predicts optimal (or near-optimal) solutions as a function of system inputs, enabling rapid inference suitable for real-time and embedded decision-making. This paradigm is particularly attractive for problems such as AC Optimal Power Flow (AC-OPF)~\cite{Zamzam2020,Park2024}, where repeated decisions of optimal generation outputs and setting points are required across time-varying grid conditions.

However, realistic AC-OPF problems are highly constrained and strongly nonconvex. They include (i) nonlinear AC power balance equations and (ii) operational inequality limits on voltages, generations, and transmission lines. These constraints lead to nonconvex parametric nonlinear programs (pNLPs) that are, in general, NP-hard. Directly learning parametric solution maps for such problems is therefore difficult, due to the rugged optimization landscape, where simple penalty-based training often converges to infeasible or poor-quality solutions.

Earlier L2O approaches predominantly followed a supervised learning paradigm~\cite{NEURIPS2021_d1942a3a}, where models are trained to mimic the solver’s outputs in labeled data. Yet, these methods inherit the limitations of their offline training data and lack explicit enforcement of feasibility. As a result, their generalization to unseen operating conditions is often weak, producing solutions that violate power flow or operational constraints.

\begin{figure}[h]
\vspace{-3mm}
	\centering
	\includegraphics[width=0.8\linewidth]{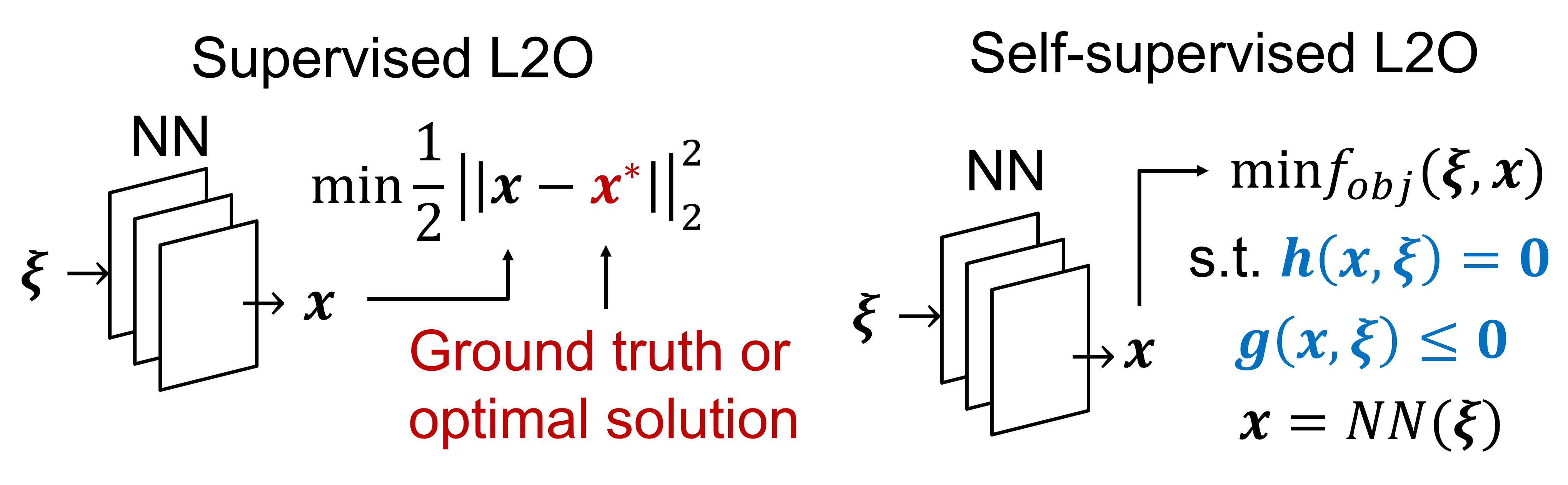}
	\caption[]{Supervised vs self-supervised learning to optimize (L2O).}
	\label{fig: imitation learning VS DPP}
    \vspace{-3mm}
\end{figure}

To address feasibility and convergence challenges, recent work has explored self-supervised L2O, where the learning objective is defined directly by the AC-OPF problem itself~\cite{NELLIKKATH2022108412,Huang2024}. This removes dependence on labeled datasets but introduces the difficulty of solving a nonconvex constrained optimization problem in the parameter space of the problem.
As a result, numerous constraint-handling strategies have been adopted, including penalty-based formulations \cite{ppf-dnn,pf-dnn-topo}, primal-dual Lagrangian training \cite{DL-lagrangian1,Fioretto_2020,Zhang2022,Pan2023}, sensitivity-informed training~\cite{Singh2022}, projected or correction-based methods \cite{DC3, Krylov, dcopf-dnn}, and completion-layer architectures \cite{DC3, ACnet}. While each method offers advantages, they can still suffer from unstable convergence in highly nonconvex settings, particularly when feasible solutions lie in narrow basins of attraction.

Broadly speaking, these new methods build on
sensitivity analysis developed in the context of operations research~\cite{mpLP76,gal_advances_1997} and later adopted in control theory applications~\cite{Bemporad2000, MPT3}. 
In the literature, these methods can be found under different names, such as constrained deep learning, optimization learning, or deep declarative networks~\cite{vanhentenryck2025,gould2021deep,kotary2021end}.

\noindent\textbf{Contributions.}
This work contributes to these efforts by introducing a homotopy-guided training approach designed to improve convergence stability in self-supervised L2O for AC-OPF.
Specifically, we make the following contributions:
\begin{enumerate}
    \item \textbf{Homotopy Continuation Heuristics for Self-Supervised L2O.}
    We propose general homotopy-based heuristics that can be integrated into any self-supervised L2O framework for constrained nonlinear optimization. Two complementary strategies are developed:  
    (a) \emph{Relaxation-based homotopy}, which gradually tightens objective weights and constraint bounds to enlarge the basin of attraction; and  
    (b) \emph{AC-OPF-aware homotopy}, which reparameterizes the problem to expose feasible starting points and maintain feasibility during training.

    \item \textbf{Empirical Validation on AC-OPF Benchmarks.}
    We evaluate the proposed approach on standard  IEEE test systems. Results show that homotopy-guided L2O substantially improves feasibility rates, while achieving objective values comparable to full AC-OPF solvers and outperforming non-homotopy L2O baselines.
\end{enumerate}

\section{Related Work}
\label{sec: related work}

Homotopy method is a type of meta-heuristics to handle hard problems, which can otherwise easily diverge or converge to a bad point. It decomposes the original (nonlinear) problem $F(x)$ into a series of sub-problems $H(x,\lambda_H)$, creating a path of optimizers driven by the change of homotopy parameter $\lambda_H$. As $\lambda_H$ shifts from $0$ to $1$, the sub-problem $H(x,\lambda_H)$ continuously transforms from a simple problem to the original one. Commonly, homotopy-based heuristics have been used for local optimization of nonlinear problems. The most popular form of designing $H(x,\lambda_H)$ is via a linear combination of a trivial problem $H_0(x)$ and the original one, such that $H(x,\lambda_H)=(1-\lambda_H)H_0(x)+\lambda_H F(x)$. Existing works have developed perturbation techniques for general nonlinear~\cite{homotopy-techniques, homotopy-nlp} and multi-objective problems~\cite{homo-multiobj}. 

Domain-specific homotopy heuristics have also been developed to design $H(x,\lambda_H)$ whose solutions are trivial when $\lambda_H=0$. For example, in the domain of circuit simulation \cite{circuit-simulation}, Gmin-stepping initially shorts all nodes to ground and gradually removes the short-circuit effect; Tx-stepping initially shorts all transmission lines, and then gradually returns to the original branches. Works in \cite{sugar-tx-stepping}\cite{sugar-imb}\cite{sugar-gmin-stepping} \cite{sugar-imb} further applied the circuit-theoretic homotopy ideas to power grid simulation and optimization tasks.
Others~\cite{homo-global}  extended homotopy methods to global optimization tasks through an ensemble of solution points, with probabilistic convergence bounds.
In this work, we build on these foundations and bring the idea of homotopy heuristics to self-supervised L2O.

\section{AC-OPF Problem Formulation}\label{sec: opf}
 
The problem of interest in this paper is the AC optimal power flow (ACOPF) \cite{acopf-review,sugar-imb}, which determines a cost-optimal generator dispatch that satisfies AC power balance and operational limits. An ACOPF formulation is:
\begin{subequations}
\label{problem: acopf}
\begin{align}
&\min_{x=[V_i,P_g,Q_g]} \sum_{g\in \mathcal{N}_g} \alpha_g P_{g}^2 + \beta_g P_{g} + \gamma_g \\
\text{s.t.}&
 (P_g - P_d) + j(Q_g - Q_d) = V \odot (Y_{\text{bus}} V)^* \label{eq: power balance}\\
& V^{\text{imag}}_{\text{slack bus}} = 0 \label{eq: ref angle}\\
& |V|_{i}^- \le |V|_{i} \le |V|_{i}^+,~~ \forall \text{ bus }i \label{ineq: Vmag bounds}\\
& P_{g}^- \le P_{g} \le P_{g}^+, \forall \text{ generator }g \label{ineq: Pg bounds}\\
& Q_{g}^- \le Q_{g} \le Q_{g}^+, \forall \text{ generator }g \label{ineq: Qg bounds}
\end{align}
\end{subequations}
where $P_d,Q_d$ are demand at each load $d$, $(\alpha_g,\beta_g,\gamma_g)$ are generator cost coefficients, $P_{g},Q_{g}$ are generator outputs, $V = V^{\text{real}} + j V^{\text{imag}}$ is the bus-voltage vector, $|V|$ its magnitude, and superscripts $-,+$ denote bounds.

\section{Homotopy Heuristics for Self-Supervised Learning to Optimize}

We propose a homotopy-based training framework for self-supervised learning to optimize (L2O) with structured transformations of objectives and constraints. A neural network policy $\pi_{\Theta}$ approximates the solution map of a constrained optimization problem, and homotopy is used to guide training from relaxed, easy-to-satisfy problems toward the original problem, improving feasibility and convergence.

\subsection{Self-Supervised Learning to Optimize}

To avoid reliance on labeled optimal solutions and address scalability in parametric nonlinear programs, we adopt a self-supervised L2O formulation  as the original problem $F(x)$:
\begin{subequations}
\begin{align}
    \min_{\Theta}  f_{\text{obj}}(x,\xi)
    \ & \text{s.t.}\ 
    g(x,\xi)\leq 0,\ 
    h(x,\xi)=0,\  \\
   & x = \pi_{\Theta}(\xi),\ \forall \xi\in\Xi,
    \label{problem: DPP}
\end{align}
\end{subequations}
where $\pi_{\Theta}$ maps problem parameters $\xi$ to decision variables $x$.

In this work, we focus on applying a penalty-based formulation for training $\pi_{\Theta}$, where the training loss function is:
\begin{equation}
\label{problem: penalty method}
    \min_{\Theta}\; 
    f_{\text{obj}}(\cdot)
    + \sum_i w_i p_i\!\left(h_i(\cdot)\right)
    + \sum_i w_i p_i\!\left(g_i(\cdot)\right)
\end{equation}
where $p_i(\cdot)$ penalize constraint violations (e.g., squared residuals for equalities and ReLU for inequalities), and gradients are computed via automatic differentiation through the neural policy $x =\pi_{\Theta}(\xi)$.

\subsection{Homotopy-Based Training}

\begin{figure}[h]
\vspace{-2mm}
	\centering
	\includegraphics[width=0.8\linewidth]{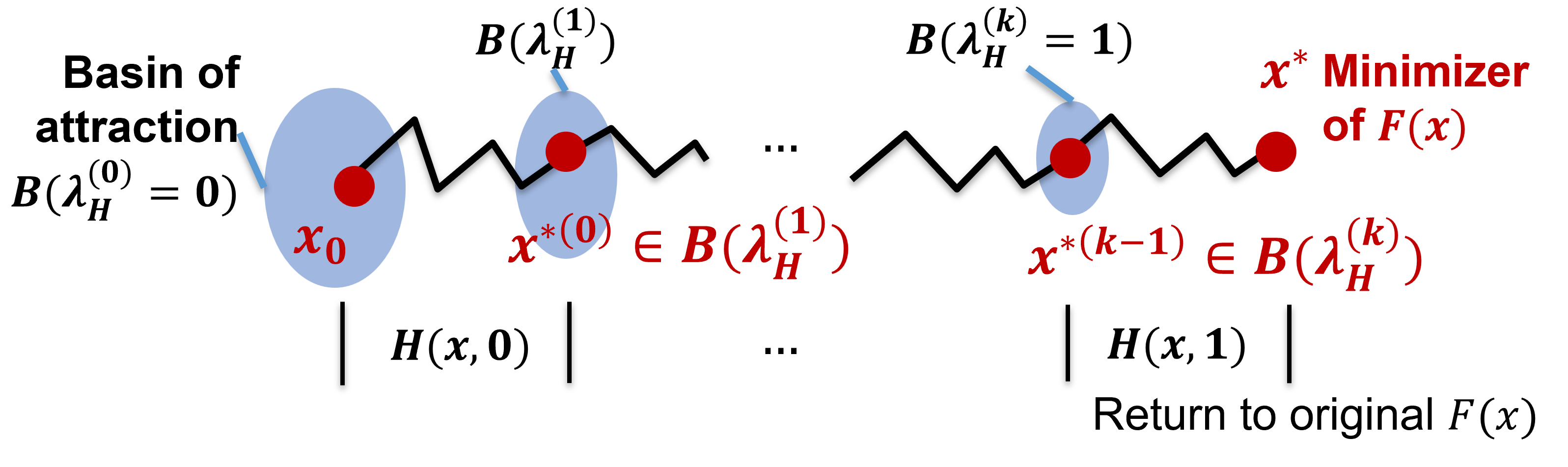}
	\caption[]{A desirable homotopy path where the previous problem solution falls within the basin of attraction for the next.}
	\label{fig: homotopy path}
    \vspace{-2mm}
\end{figure}

Direct optimization of \eqref{problem: penalty method} for the original highly nonconvex, constrained problem often suffers from infeasible solutions. We therefore introduce a homotopy parameter $\lambda_H \in [0,1]$ and train $\pi_{\Theta}$ over a sequence of homotopy problems $H(x,\lambda_H)$:
\begin{equation}
\min_{\Theta}  f_{\lambda_H}(\cdot)
    \text{ s.t.}\ 
    g_{\lambda_H}(\cdot)\leq 0,\ 
    h_{\lambda_H}(\cdot)=0, \text{(\ref{problem: DPP})} 
    \label{problem: homotopy DPP}
\end{equation}
where $(f_{\lambda_H}, g_{\lambda_H}, h_{\lambda_H})$ interpolate between relaxed and original formulations. The same penalty method as in (\ref{problem: penalty method}) is applied to the gradually tightened problem $H(x,\lambda_H)$, leading to
\begin{equation}
\label{problem: homotopy}
    \min_{\Theta}
    f_{\lambda_H}(\cdot)
    + \sum_i w_i p_i\!\left(h_{\lambda_H,i}(\cdot)\right)
    + \sum_i w_i p_i\!\left(g_{\lambda_H,i}(\cdot)\right)
    \vspace{-1mm}
\end{equation}
Training proceeds by:
(i) initializing at an easy problem with $\lambda_H=0$,
(ii) optimizing $\Theta$ for a fixed number of steps,
(iii) incrementing $\lambda_H \leftarrow \lambda_H + \Delta\lambda_H$,
and (iv) continuing from the previous $\Theta$ as a warm start.
This schedule gradually approaches feasible regions of the original hard problem instead of learning it from scratch, as in Fig. \ref{fig: homotopy path}.
We then design two complementary types of homotopy transformations.
\subsection{Type I: Relaxation-Based Homotopy}
\label{sec: heuristics type 1}

Type I heuristics start from a relaxed objective and constraints set with a large feasible basin, and then gradually enforce the true problem.

\paragraph*{Convexify Objective (CObj)}
Replace a nonconvex $f_{\text{obj}}$ by a convex surrogate $f_{\text{cvx}}$ at $\lambda_H=0$ and interpolate:
\begin{equation}
    f_{\lambda_H} = (1-\lambda_H) f_{\text{cvx}} + \lambda_H f_{\text{obj}}.
\end{equation}

\paragraph*{Shrink Bounds (SBnds)}
Start with loose initial bounds $x\in[\epsilon^{-},\epsilon^{+}]$ and gradually tighten them back to $x\in[x^{-},x^{+}]$:
\begin{equation}
    x^{-}_{\lambda_H} = (1-\lambda_H)\epsilon^{-} + \lambda_H x^{-};
    x^{+}_{\lambda_H} = (1-\lambda_H)\epsilon^{+} + \lambda_H x^{+}
\end{equation}

\paragraph*{Grow Penalty (GPen)}
Scale (inequality) constraint penalties with $\lambda_H$ so that violations are weakly penalized early and strictly penalized later:
\begin{equation}
   \lambda_H g(\cdot)\leq0
\end{equation}

\paragraph*{Split and Shrink (SaS) for Equalities}
Convert $h(\cdot)=0$ to inequalities $-\epsilon_{\lambda_H} \leq h(\cdot)\leq \epsilon_{\lambda_H}$ with
\begin{equation}
    \epsilon_{\lambda_H} = (1-\lambda_H)\epsilon_H + \lambda_H \epsilon_L,
\end{equation}
where $\epsilon_H \gg \epsilon_L \approx 0$. As $\lambda_H\to 1$, equalities are enforced tightly.

We combine CObj, SBnds, SaS, and GPen into \eqref{problem: homotopy} to obtain a smooth transformation from relaxed to strict constraints.

\subsection{Type II: AC-OPF-Aware Homotopy with Trivial Solutions}
\label{sec: heuristics type 2}

Type II heuristics exploit power-system structures to construct homotopies for problems like ACOPF.

\paragraph*{Load-Stepping}
We gradually increase loads from zero to their original values:
\begin{equation}
    h_{\lambda_H}:
    \ (P_g - \lambda_H P_d) + j(Q_g - \lambda_H Q_d)
    - v \odot (Y_{\text{bus}} v)^* = 0.
\end{equation}
At $\lambda_H=0$, a trivial no-load solution $x_0$ exists (intuitively, voltage and power outputs should be all zeros when loads are zero); the network is first trained to satisfy this easy condition (optionally regularized toward some labels $x_0$), and then moves back to the original load condition as $\lambda_H$ increases.

\paragraph*{Tx-Stepping}
We deform the system admittance matrix:
\begin{equation}
    Y_{\lambda_H} = (1-\lambda_H) Y_0 + \lambda_H Y_{\text{bus}},
\end{equation}
where $Y_0$ corresponds to low-impedance (nearly shorted) lines. At $\lambda_H=0$, the system is well-conditioned with voltages close to the reference at all buses, yielding an easy feasible solution. As $\lambda_H$ grows, the NN is trained to maintain feasibility under the gradually restored physical network.

Both load-stepping and Tx-stepping can be combined with Type I relaxations inside the same homotopy schedule, enabling a robust training procedure for generic L2O.

\section{Numerical Results}
\label{sec:Results}
We evaluate the efficacy of homotopy heuristics on both the power grid optimal power flow problem and general random non-convex constrained optimization problems to assess generality. 
The use of homotopy is expected to enable a more reliable convergence of the neural network models to outperform non-homotopy results on the following criteria:
\begin{itemize}
    \item \textbf{Optimality}: the objective function $f_{obj}$ achieved by the solution. For the ACOPF problem, it represents the per-hour cost (\$/h) of the generation dispatch.
    \item \textbf{Feasibility}: how much the solution $\hat{x}$ violates the equality and inequality constraints. Feasibility is quantified by the mean and maximum violation of constraints: $\text{mean}(h(\hat{x}))$, $\max(h(\hat{x}))$, $\text{mean}(relu(g(\hat{x})))$, $\max(relu(g(\hat{x})))$. In real-world tasks, smaller violation of constraints means the neural network outputs a more practical solution for real-world optimization and control.
\end{itemize}

We compare the different versions of homotopy L2O with a non-homotopy baseline represented by the penalty method~(\ref{problem: penalty method}).  Appendix \ref{sec: appendix} describes the details on experiment settings and hyper-parameter tuning, for a fair comparison.



\subsection{Power Grid AC Optimal Power Flow Problem}
This subsection evaluates the real-world task of the ACOPF problem (defined in Section \ref{sec: opf}). 
Table \ref{tab: penalty, case30, acopf} shows ACOPF results on a 30-bus system. Homotopy methods improve the feasibility of NN outputs with smaller violations of equality and inequality constraints. See Appendix \ref{sec: appendix} for our experiment settings and hyper-parameter tuning.

\setlength{\tabcolsep}{6pt}
\begin{table*}[htbp]
\small
\centering
	\begin{tabular}{ @{}r|ccccc@{} }  
	\toprule
	\textbf{Method} & \textbf{Obj} &\textbf{Mean eq.} & \textbf{Max eq.} & \textbf{Mean ineq.} & \textbf{Max ineq.} \\ 
	\midrule
\textbf{SaS, SBnds}& 
 666 & 0.0027 (0.0015) & 0.010 (0.005) & 0.0000 (0.0000) &0.000 (0.001)\\
\textbf{SaS, GPen}& 666 & 0.0023 (0.0014) & 0.008 (0.005) & 0.0000 (0.0000) & 0.000 (0.000)\\
\textbf{Tx-stepping, SBnds}& 665 & 0.0024 (0.0013) & 0.008 (0.005) & 0.0000 (0.0000) & 0.000 (0.000)\\
\textbf{Tx-stepping, GPen}& 665 & 0.0017 (0.0009) &0.006 (0.003) & 0.0000 (0.0000) & 0.000 (0.000)\\
\textbf{Load-stepping, SBnds}& 666 & 0.0020 (0.0012) & 0.007 (0.004) & 0.0000 (0.0000) & 0.000 (0.000)\\
\textbf{Load-stepping, GPen}& 667 & 0.0024 (0.0016) & 0.009 (0.005) & 0.0000 (0.0000) & 0.000 (0.000)\\
\textbf{Original penalty}& 673 & 0.0036 (0.0027) & {\color{black}0.013 (0.010)} & 0.0000 (0.0000) & {\color{black}0.001 (0.003)}\\
	\bottomrule
	\end{tabular} 
	\caption{Results of ACOPF problem on case30, over 100 test instances. Mean and max violations are analyzed across instances, we list the \textsf{average value (std)}. The original penalty method has larger violations of constraints, whereas homotopy methods have smaller violations. 
	\label{tab: penalty, case30, acopf}}
    \vspace{-1mm}
\end{table*}

\subsection{Non-convex Optimization with Random System Constraints}\label{sec: lincon experiments}
First, consider a problem with a non-convex objective and random linear inequality constraints:
\begin{equation}
   \min_x \sum_{i=1}^{n-1}(1-x_i)^2 +  2(x_{i+1}-x_i^2)^2 
   \hskip 0.5em \text{s.t. } \hskip 0.5em Ax\leq b + C\xi
   \label{problem: noncvx obj, lincon}
\end{equation}
where
$n$ is the problem size (complexity), $x$ is the solution vector representing $n$ variables to solve, $\xi$ is an $round(0.4*n)\times 1$ vector representing the (known) input parameter that varies across instances, $A^{n\times n}, b^{n\times 1}, C^{dim(\xi)}$ are randomly generated matrices representing $n$ random linear constraints.

Table \ref{tab: penalty, lincon} shows results on test data for problems with varying complexity. Results demonstrate that adding homotopy heuristics onto the penalty method enables neural network outputs to have a smaller violation of constraints. See Appendix \ref{sec: appendix} for details on experiment settings.

\setlength{\tabcolsep}{3pt}
\begin{table*}[htbp]
\small
\centering
	\begin{tabular}{ @{}r|lcccc@{} }  
	\toprule
	\textbf{Method} && & \textbf{Complexity $n$} & & \\ 
	\textbf{} &&\begin{tabular}{c}
	      5\\ 
	     $\Delta\lambda_H=0.05$
	\end{tabular} & 
	\begin{tabular}{c}
	     25 \\ $\Delta\lambda_H=0.05$
	\end{tabular}
	& \begin{tabular}{c}
	       50 \\ $\Delta\lambda_H=0.05$
	\end{tabular} 
	& \begin{tabular}{c}
	     100 \\ $\Delta\lambda_H=0.005$, $w_{ineq}=500$
	\end{tabular}  \\ 
	\midrule
\textbf{CObj, SaS, SBnds}&
\begin{tabular}{r}
     obj\\
     mean viol\\
     max viol
\end{tabular}&
\begin{tabular}{l}
     0.48 \\  0.0000(0.0000)\\ 0.0000(0.0000)
\end{tabular}
& \begin{tabular}{l}
     31.71 \\0.0001(0.0002) \\0.0016(0.0048)
\end{tabular}&
\begin{tabular}{l}
62.03 \\ 0.0003(0.0011) \\0.0129(0.0526)
\end{tabular}
& \begin{tabular}{l}
  276.16 \\0.0002(0.0009) \\0.0246(0.0947)
\end{tabular}\\ 
\midrule
\textbf{CObj, SaS, GPen}&
\begin{tabular}{r}
     obj\\
     mean viol\\
     max viol
\end{tabular}& 
\begin{tabular}{l}
     0.48 \\0.0000(0.0000) \\0.0000(0.0000)
\end{tabular}
& 
\begin{tabular}{l}
     31.69 \\0.0001(0.0005)\\ 0.0029(0.0121)
\end{tabular}
& \begin{tabular}{l}
     62.59 \\0.0002(0.0005) \\0.0061(0.0141)
\end{tabular}
& \begin{tabular}{l}
     253.42 \\0.0004(0.0018) \\0.0297(0.1032)
\end{tabular}\\ 
\midrule
\textbf{SaS, SBnds}&
\begin{tabular}{r}
     obj\\
     mean viol\\
     max viol
\end{tabular}&
\begin{tabular}{l}
     0.48 \\ 0.0000(0.0000) \\ 0.0000(0.0000)
\end{tabular}
&
\begin{tabular}{l}
     33.35 \\0.0001(0.0003) \\0.0023(0.0066)
\end{tabular}
&\begin{tabular}{l}
     62.33 \\0.0002(0.0009) \\0.0082(0.0222)
\end{tabular} 
& \begin{tabular}{l}
     272.36 \\0.0002(0.0009) \\0.0180(0.0711)
\end{tabular} \\ 
\midrule
\textbf{SaS, GPen}&
\begin{tabular}{r}
     obj\\
     mean viol\\
     max viol
\end{tabular}& \begin{tabular}{l}
      0.47 \\ 0.0000(0.0000)\\ 0.0000(0.0000)
\end{tabular}
& \begin{tabular}{l}
     32.15 \\0.0001(0.0004)\\0.0024(0.0080)
\end{tabular}
& 
\begin{tabular}{l}
     62.20 \\0.0006(0.0035) \\0.0095(0.0347)
\end{tabular}
& \begin{tabular}{l}
     447.32 \\{\color{black}0.0031(0.0067)} \\{\color{black}0.1990(0.3583)}
\end{tabular}\\ 
\midrule
\begin{tabular}{r}
\textbf{Original} \\
(penalty method)
\end{tabular}&
\begin{tabular}{r}
     obj\\
     mean viol\\
     max viol
\end{tabular}& \begin{tabular}{l}
     0.46 \\0.0000(0.0000)  \\ 0.0000(0.0000)
\end{tabular}
&\begin{tabular}{l}
   31.84 \\0.0005(0.0010) \\{\color{black}0.0095(0.0160)}
\end{tabular} 
&\begin{tabular}{l}
     62.44\\ 0.0005(0.0011)\\ {\color{black}0.0193(0.0470)}
\end{tabular} 
& \begin{tabular}{l}
     208.71 \\ {\color{black}0.0015(0.0023)} \\ {\color{black}0.1249(0.1877)}
\end{tabular} \\ 
	\bottomrule
	\end{tabular} 
 \caption{Non-convex problem with $n$ variables and $n$ random linear constraints as $n$ varies as $5, 25, 50, 100$. Results over 100 test instances are listed, using metrics of objective, mean, and max inequality constraint violations. The violations are formatted as \textsf{average value (std)} in this Table. Results show that the homotopy heuristics enable a smaller violations than the original penalty method.}
 	\label{tab: penalty, lincon}
    \vspace{-3mm}
\end{table*}

\section{Conclusion}
This work introduced a homotopy-guided training heuristic for self-supervised Learning to Optimize (L2O) in highly nonconvex constrained optimization problems, with a focus on AC Optimal Power Flow. The proposed homotopy heuristics reshape the optimization landscape by (i) gradually relaxing and tightening objectives and constraint bounds to expand and guide the basin of attraction, and (ii) exploiting power-system structure to construct domain-aware continuation paths with trivial feasible starting points. These mechanisms are general and can be integrated into a broad range of L2O methods.

Numerical experiments on AC-OPF test cases and synthetic nonconvex constrained problems show that, compared to the penalty-based approach, homotopy-guided training consistently improves convergence reliability and significantly reduces constraint violations, while maintaining near-optimal objective performance. These results highlight homotopy continuation as an effective heuristic for improving the convergence of self-supervised L2O in power system applications.

\section*{Acknowledgments}

This research was partially supported by the Data Model Convergence (DMC) initiative at Pacific Northwest National Laboratory (PNNL), 
and by the Ralph O’Connor Sustainable Energy Institute at Johns Hopkins University.

\appendix
\section{Appendix}\label{sec: appendix}

\subsection{Experiment settings and hyper-parameter tuning}

To create a fair comparison in each optimization problem, experiments with and without homotopy heuristics will train with the same NN architecture, Adam optimizer, and learning rate scheduler (\textsf{StepLR} with step=100, gamma=0.1, and a minimal learning rate $10^{-5}$; in homotopy methods, lr scheduler is only applied to the last homotopy step). 
Early stopping is also applied in each experiment to avoid overfitting: each original penalty method trains for 1,000 epochs with {warmup=50, patience=200}, and each homotopy method trains 100 epochs in each homotopy step with warmup=50, patience=50.  
All NNs are trained on \textsf{PyTorch}.

Experiment settings for the non-convex problem with random linear constraints, see problem definition (\ref{problem: noncvx obj, lincon}) in Section \ref{sec: lincon experiments}, are listed below:
\begin{itemize}
    \item Dataset: 50,000 instances (with train/validation/test ratio 8:1:1)
    \item NN architecture: cylinder NN with 4 layers, hidden layer size increases with problem size $n$ by $hidden layer size = 30\sqrt{n/5}$
    \item Penalty weights: $w_{eq}=50$, $w_{ineq}=50$
    \item Homotopy settings: CObj has $f_{cvx}= \sum_{i=1}^{n-1}(1-x_i)^2 +  2(x_{i+1}^2+x_i^4) $, SaS has $\epsilon_H=0.01, \epsilon_L=0$, SBnds has $\epsilon^+=1, \epsilon^-=0$.
\end{itemize}
These hyper-parameters are tuned and kept fixed across all experiments of this non-convex problem.

For the ACOPF problem, we propose an additional trick of $P_g$ pull-up and used it on all experiments. 

\noindent\textbf{($P_g$ pull-up)} Based on power system domain knowledge, any decision with supply lower than demand is always technically infeasible. To avoid bad predictions of this type, we apply the heuristics of $P_g$ pull-up where an additional domain-specific constraint $\sum P_g - \sum P_d \geq \epsilon$ is added to pull the generation up and thus promote convergence to a point with total supply higher than demand. This constraint is not subject to homotopy heuristics and remains the same in the homotopy path. experiment settings are as follows. Similarly as for other constraints, we have a weight $w_{pullup}$ to impose the additional constraint in penalty method.

\begin{itemize}
    \item Dataset: 50,000 instances (with train/validation/test ratio 8:1:1), data are generated by randomly sampling load profiles in the range of $75\%-150\%$ of the base load profile (base load is the load in case data). 
    \item Batchsize: 1024
    \item NN architecture: cylinder NN with 2 layers and hidden layer size = 200,
    \item Penalty weights: $w_{eq}=10^{5}$, $w_{ineq}=10^{6}$
    \item General homotopy settings: $\Delta\lambda_H=0.05$, SaS has $\epsilon_H=0.01, \epsilon_L=0$, SBnds has $\epsilon^+=1, \epsilon^-=0$.
    \item Domain specific homotopy settings: warm homotopy loss has $w_{warm}=10^6$ 
    \item others: $P_g$ pull up has  $w_{pullup}=10^6, \epsilon=0.01$
\end{itemize}
The hyper-parameters are kept fixed across all experiments.


\bibliographystyle{IEEEtran}
\bibliography{main}

\end{document}